\documentclass[letterpaper]{article} 
\usepackage{aaai2027}  
\usepackage[hyphens]{url}  
\usepackage{graphicx} 
\usepackage{natbib}  
\usepackage{caption} 
\usepackage{algorithm}
\usepackage{amsmath,amssymb}
\usepackage{algpseudocode}
\usepackage{algorithmicx}
\usepackage{multirow} 

\usepackage{newfloat}
\usepackage{listings}
\DeclareCaptionStyle{ruled}{labelfont=normalfont,labelsep=colon,strut=off} 
\floatstyle{ruled}
\newfloat{listing}{tb}{lst}{}
\floatname{listing}{Listing}

\usepackage{booktabs}

\title{Rethink Before You Execute: Adaptive Execution for World Action Models
}
\author{
    Feng Ye\textsuperscript{\rm 1}, 
    Yiming Zhao\textsuperscript{\rm 2}, Yong Yu\textsuperscript{\rm 2}, Hongxu Zhou\textsuperscript{\rm 2}, Yong Pan\textsuperscript{\rm 2}, \\
    Yuan Xue\textsuperscript{\rm 1}, Peng Jia\textsuperscript{\rm 2}, Chuanmin Jia\textsuperscript{\rm 1}\corresponding
}
\affiliations{
    \textsuperscript{\rm 1}Wangxuan Institute of Computer Technology, Peking University, 
    \textsuperscript{\rm 2}Simplexity Robotics\\

}

\begin{document}

\nocopyright
\maketitle

\begin{abstract}
World Action Models (WAMs) jointly predict future actions and the evolution of the environment. At each inference, a WAM generates a chunk of actions and the robot executes a fixed prefix before replanning. 
We argue that this fixed execution horizon is poorly matched to execution dynamics: the chunk reliability varies across task stages, so when to replan depends on the result of accumulated execution, not on the step counts. 
We propose TempoWAM (\textbf{T}iming \textbf{E}xecution by \textbf{M}onitoring \textbf{P}rogress \textbf{O}nline), a lightweight plug-and-play execution scheme for WAMs. A Recurrent Progress Monitor first estimates task progress from the current observation, task instruction, remaining actions, and execution history; and an Adaptive Execution Protocol then evaluates whether the chunk is advancing the task to decide if replanning is needed.
To bridge the training–deployment gap, the protocol is calibrated by a task-dependent calibration factor with online adaptation. 
Experiments on LIBERO, RoboTwin, and real-world tasks show that TempoWAM consistently improves the efficiency-success trade-off of WAM execution. On real robots, it reduces WAM inferences by 26.9\% on easy tasks while maintaining success, and improves success by 13.3 points on difficult tasks.
\end{abstract}


\section{Introduction}

\begin{figure}[t]
    \centering
    \includegraphics[width=\linewidth]{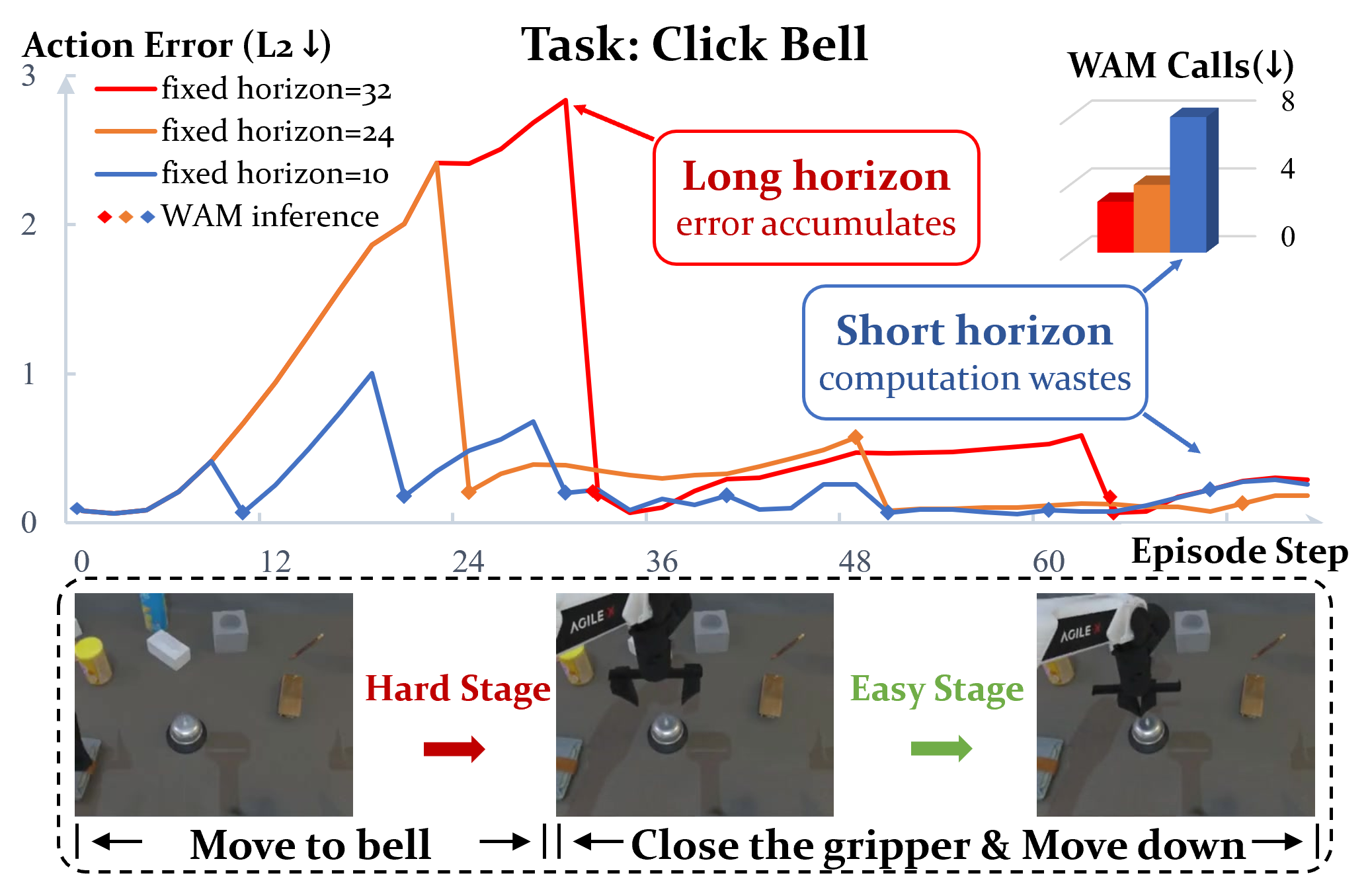}
    \caption{\textbf{Stage-dependent chunk reliability in Click Bell task of RoboTwin 2.0.} We plot the L2 error of action under three fixed execution horizons. Chunk reliability varies across execution stages: a long horizon accumulates error in hard stages, while a short horizon remains accurate but triggers unnecessary WAM calls in easy stages where the chunk is still reliable. This mismatch shows that no single fixed horizon fits all stages, motivating adaptive execution.}
    \label{fig:motivation}
\end{figure}

World Action Models (WAMs) have emerged as a powerful paradigm for robot manipulation, jointly predicting future actions and the evolution of the environment in a single model. At each inference step, a WAM predicts a chunk of future actions, and the robot then executes a fixed prefix of the chunk before replanning. 
However, as shown in Fig.~\ref{fig:motivation}, the reliability of chunks is not uniform across the task: during easy stages, a chunk can remain accurate for many steps; but during hard stages, errors can accumulate rapidly after only a few steps. 
This suggests that when to stop and replan should depend on the execution state, not on a predetermined number of executed steps. 

Existing works have begun to explore adaptive execution, but most rely on proxy metrics such as action entropy, prediction consistency, or attention weights.  
These metrics do not explicitly measure what actually matters for execution: whether the task is advancing. An action sequence can be smooth yet make no progress, and a high-entropy one may still be advancing the task. 
For WAMs, this distinction is especially important: they model both actions and environment evolution, so the execution decision should be made based on whether the remaining actions are still advancing the task, not merely on whether the chunk looks plausible.


\begin{figure*}[t]
    \centering
    \includegraphics[width=0.95\linewidth]{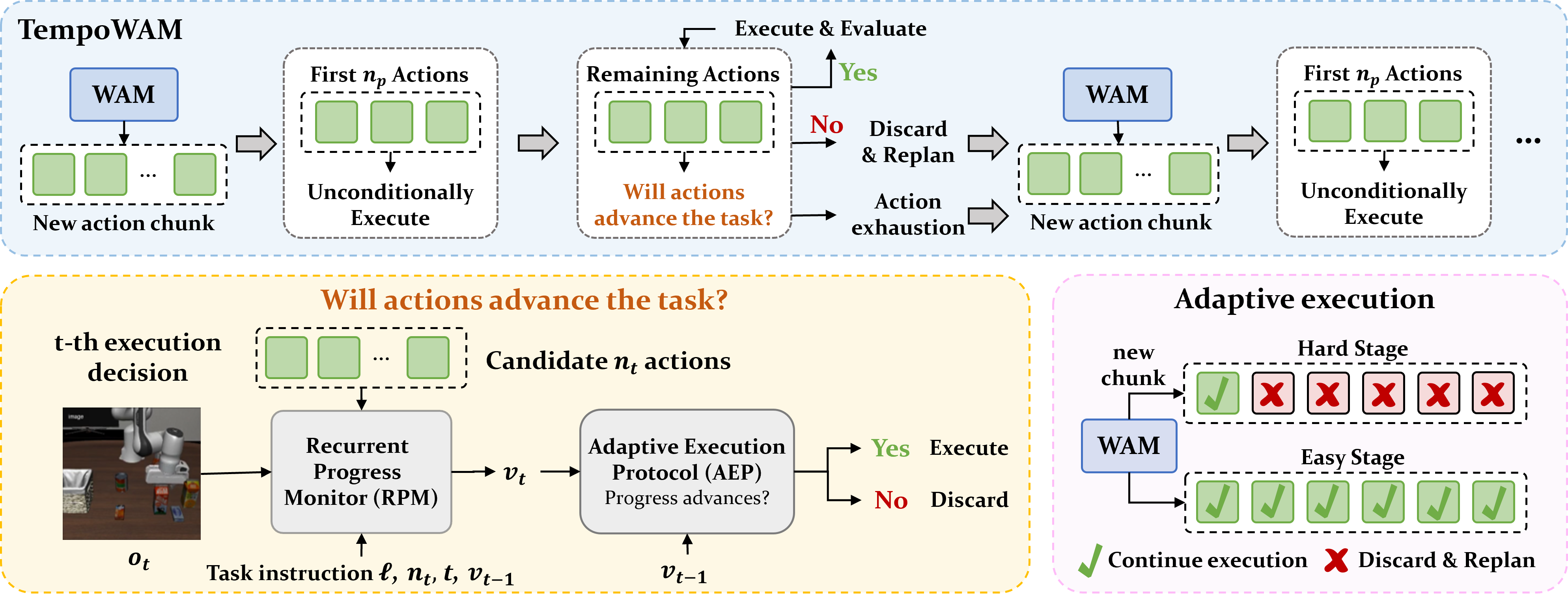}
    \caption{\textbf{Overview of TempoWAM.} The base WAM generates an action chunk $c$ from the current observation, whose first $n_p$ actions are executed unconditionally. Each subsequent decision step evaluates the remaining $n_t=\min(n_p,|c^{rem}|)$ candidate actions: the RPM estimates the task progress that would be reached if they were executed, from the current observation, task instruction, candidate prefix, and execution history; the AEP then decides whether they will advance the task at the required rate. If yes, execution continues; otherwise the remaining actions are discarded and the WAM is triggered for a fresh chunk. TempoWAM adapts the execution horizon online: easy stages run through with few replanning, while hard stages trigger early discard and frequent replanning (bottom right).
    }
    \label{fig:motivation1}
\end{figure*}

Based on this observation, we propose \textbf{TempoWAM}, a lightweight adaptive execution scheme for WAMs. TempoWAM introduces a \textbf{Recurrent Progress Monitor (RPM)} that estimates task progress from the current observation, task instruction, remaining actions, and execution history. Given this estimate, an \textbf{Adaptive Execution Protocol (AEP)} decides whether to continue executing or to discard the remaining actions and replan. To bridge the gap between training supervision and actual execution, we further introduce a simple per-task calibration factor with online adaptation.

This design gives TempoWAM three desirable properties. First, it is \textbf{plug-and-play}: it achieves adaptive execution without modifying or retraining the backbone WAM. Second, it is \textbf{progress-aware}: the execution decision is driven by task advancement rather than indirect uncertainty proxies. Third, it is \textbf{efficient}: the monitor is lightweight and adds negligible overhead compared to a WAM inference call. 
Experiments on LIBERO, RoboTwin, and real-world tasks show that TempoWAM consistently improves the efficiency-success trade-off: it reduces unnecessary replanning on easy tasks and triggering earlier replanning on difficult tasks. 

Our main contributions are as follows:
\begin{enumerate}
    \item We show that assessing the rationality of actions alone is insufficient for adaptive execution, and argue that task progress is a natural criterion for this decision.
    \item We propose TempoWAM, a plug-and-play adaptive execution scheme. By grounding execution decisions in estimated task progress, TempoWAM adapts execution horizon to actual state: continuing execution when reliable and replanning when errors accumulate.
    \item We show that TempoWAM consistently improves the efficiency-success trade-off on WAM benchmarks and real robot. On real-world tasks, it reduces WAM inferences by 26.9\% on easy tasks while maintaining success, and improves success by 13.3 points on difficult tasks.
\end{enumerate}
      
\section{Related Work}
\label{sec:related}

\paragraph{World action models.}
WAMs couple future prediction and action generation within a unified policy for robot manipulation. Early approaches~\cite{du2023advances} formulate robot control through future video generation followed by action decoding, while more recent systems~\cite{hu2025video, kim2026cosmos, bi2025motus} integrate future imagination and action prediction together. 
A parallel line of work reduces the deployment cost: FastWAM~\cite{yuan2026fastwam} questions whether test-time future imagination is needed, AHA-WAM~\cite{cai2026ahawam} amortizes world planning across multiple action updates through asynchronous execution, and LaWAM~\cite{chen2026lawam} exposes predictive dynamics through compact latent visual subgoals instead of future video. Across this spectrum, research has focused on what a WAM should generate and how efficiently it can do so. 

\paragraph{Adaptive execution.}
Most robot policies execute action chunks under a fixed execution horizon. Analyses have shown that no single horizon fits all situations~\cite{liu2024bid, nie2026pace}. Recent methods adapt the horizon at inference time with four kinds of signals: uncertainty over sampled actions, via entropy~\citep{liang2026adaptive} or denoising variance~\citep{feng2026denoising}; intrinsic model signals such as action self-attention~\citep{wang2026autohorizon}; verification of imagined futures against observations~\citep{wang2026trust}; and horizon branches learned via online RL~\citep{zhao2026dynamic}.
These signals are particularly costly for WAMs: sampling-based ones multiply the WAM calls that dominate deployment cost; attention-based ones ignore the environment state that WAMs explicitly model; verification requires test-time future imagination that efficient WAMs deliberately omit; and RL-based ones need costly online interaction. TempoWAM instead drives the decision with task progress estimated directly from the environment state, using a lightweight monitor and requiring no future imagination from the backbone.

\paragraph{Task progress estimation.}
Task progress has been used as an evaluation and guidance signal: ProgressVLA~\citep{yan2026progressvla} pretrains a progress estimator and injects it into diffusion sampling via classifier guidance, steering action generation toward progress-maximizing chunks.
DEHP also suggests that execution should adapt to task progress, but leaves progress implicit in RL returns. 
TempoWAM instead uses progress at the execution layer: the RPM estimates it online to decide whether candidate actions should be executed.
   
\section{Method}\label{sec:method}


\subsection{Problem Setup}\label{sec:formulation}
We consider a WAM with action chunking. At environment time step $u$, given the current observation $o_u$ and task instruction $\ell$, a WAM predicts a chunk of future actions:
\begin{equation}
c = \pi_\theta(o_u, \ell) \in \mathbb{R}^{H \times d_a},
\end{equation}
where $H$ is the chunk horizon and $d_a$ is the action dimension.
In standard execution, the robot executes a fixed prefix and then replans. While simple, it ignores that the reliability of a chunk is stage-dependent: chunks remain accurate for a long time, while others become unreliable after only a few steps.

Our goal is to replace this fixed execution rule with an adaptive one, without modifying the backbone WAM. TempoWAM operates entirely at the execution layer and decides whether the remaining actions of current chunk should still be trusted or whether a fresh WAM call should be triggered.

\subsection{Overview of TempoWAM}\label{sec:overview}
As shown in Fig.~\ref{fig:motivation1}, TempoWAM consists of two components: the former is RPM, a monitor that estimates task progress that would be reached if the current candidate actions were executed; the latter is AEP, an execution protocol that compares the estimated progress rate with a calibrated target rate and decides whether to continue executing the current chunk or to replan.

Each time the WAM produces a new chunk, TempoWAM first unconditionally executes a short prefix of length $n_p < H$, which guarantees that the environment keeps evolving. It then repeatedly considers the next $n_t=min(n_p,|c_t^{rem}|)$ actions of the remaining chunk $c_t^{rem}$ as a candidate prefix and evaluates whether executing them keeps the task advancing: if so, execution continues; otherwise, the rest of the chunk is discarded and the WAM is called again. 

\subsection{RPM Module} \label{sec:rpm}
At execution decision step $t$, the monitor takes as input the current observation $o_t$, the remaining action candidate $c^{\mathrm{rem}}_t$, the candidate prefix length $n_t$, the previous progress estimate $v_{t-1}$, and the task instruction $\ell$:
\begin{equation}
v_t = f_{\mathrm{RPM}}(o_t, c^{\mathrm{rem}}_t, n_t, v_{t-1}, t, \ell), \  v_t \in [0,1].
\end{equation}
Here, $v_t$ estimates the task progress that would be reached after executing the candidate prefix. Values near $0$ indicate early-stage execution, while values near $1$ indicate that the task is close to completion.

\paragraph{Observation encoding.} 
We encode the observation $o_t$ with the frozen visual backbone $E_{\rm obs}$ of the WAM and project it into the monitor space by a lightweight MLP:
$z_t^o = f_{\mathrm obs}(E_{\mathrm obs}(o_t))$.
The visual backbone of a WAM is usually a video-pretrained spatio-temporal encoder, producing features with rich visual representations. 

\paragraph{Action-prefix encoding.} 
To verify whether the remaining chunk is still useful, RPM explicitly inspects the candidate actions to be executed. The first $n_t$ actions of $c^{\mathrm{rem}}_t$ are encoded by a recurrent encoder, whose final hidden state serves as the prefix representation $z_t^a = f_{\mathrm act}(c_t^{\mathrm rem}[:n_t],n_t)$. An embedding of the prefix length $n_t$ is added, so that the monitor can distinguish the different length of candidate prefixes.

\paragraph{History recurrence.}
Execution decisions should depend on both the current candidate actions and how progress has evolved so far. We therefore inject the previous progress estimate $v_{t-1}$ into the monitor through a learned projection:
\begin{equation}
z^v_t =
\begin{cases}
f_v(v_{t-1}), & \text{if } v_{t-1} \text{ is available}, \\
e_0, & \text{otherwise},
\label{equ:e0}
\end{cases} 
\end{equation}
where $e_0$ is a learned start token used at the beginning of each episode. 

\begin{table*}[t]
\centering
\small
\setlength{\tabcolsep}{5pt}
\begin{tabular}{ll c ccc ccc}
\toprule
& \multirow{2}{*}{\textbf{$n$}} & \multirow{2}{*}{\textbf{Method}} & \multicolumn{3}{c}{\textbf{Clean}} & \multicolumn{3}{c}{\textbf{Random}} \\
\cmidrule(lr){4-6} \cmidrule(lr){7-9}
& & & SR & Calls & Steps & SR & Calls & Steps \\
\midrule
\multirow{3}{*}{Short} & \multirow{3}{*}{35}
  & Baseline-24 & 93.71 & 5.97 & 130.65 & 94.14 & 6.01 & 131.45 \\
  & & Baseline-12 & 93.91 {\footnotesize(+0.20)} & 11.36 {\footnotesize(+5.39)} & 129.10 {\footnotesize($-$1.55)} & \textbf{94.37} {\footnotesize(+0.23)} & 11.42 {\footnotesize(+5.41)} & 129.83 {\footnotesize($-$1.62)} \\
  & & TempoWAM & \textbf{94.74} {\footnotesize(+1.03)} & \textbf{5.51} {\footnotesize($-$0.46)} & \textbf{127.61} {\footnotesize($-$3.04)} & 94.17 {\footnotesize(+0.03)} & \textbf{5.41} {\footnotesize($-$0.60)} & \textbf{129.26} {\footnotesize($-$2.19)} \\
\midrule
\multirow{3}{*}{Long} & \multirow{3}{*}{15}
  & Baseline-24 & 86.93 & \textbf{15.02} & 346.81 & 84.47 & \textbf{15.20} & 352.06 \\
  & & Baseline-12 & 86.27 {\footnotesize($-$0.67)} & 28.67 {\footnotesize(+13.65)} & 335.94 {\footnotesize($-$10.87)} & 83.13 {\footnotesize($-$1.33)} & 29.84 {\footnotesize(+14.64)} & 350.08 {\footnotesize($-$1.98)} \\
  & & TempoWAM & \textbf{88.47} {\footnotesize(+1.54)} & 21.67 {\footnotesize(+6.65)} & \textbf{331.66} {\footnotesize($-$15.15)} & \textbf{85.33} {\footnotesize(+0.87)} & 22.24 {\footnotesize(+7.04)} & \textbf{341.04} {\footnotesize($-$11.02)} \\
\midrule
\multirow{3}{*}{Overall} & \multirow{3}{*}{50}
  & Baseline-24 & 91.68 & \textbf{8.69} & 195.49 & 91.24 & \textbf{8.76} & 197.63 \\
  & & Baseline-12 & 91.62 {\footnotesize($-$0.06)} & 16.56 {\footnotesize(+7.87)} & 191.15 {\footnotesize($-$4.34)} & 91.00 {\footnotesize($-$0.24)} & 16.94 {\footnotesize(+8.18)} & 195.91 {\footnotesize($-$1.72)} \\
  & & TempoWAM & \textbf{92.86} {\footnotesize(+1.18)} & 10.36 {\footnotesize(+1.67)} & \textbf{188.83} {\footnotesize($-$6.66)} & \textbf{91.52} {\footnotesize(+0.28)} & 10.46 {\footnotesize(+1.70)} & \textbf{192.80} {\footnotesize($-$4.83)} \\
\bottomrule
\end{tabular}
\caption{Results on RoboTwin. Baseline-24 and Baseline-12 denote FastWAM with fixed execution horizon of 24 and 12. TempoWAM consistently improves the efficiency-success trade-off in both groups: on short-horizon tasks, it preserves or improves success while reducing calls; on long-horizon tasks, it improves success while avoiding unnecessary execution.}
\label{tab:robotwin-grouped}
\end{table*}

\paragraph{Aggregation and prediction. }
The observation, action-prefix, history, and position embeddings are fused into a single recurrent state:
\begin{equation}
z_t = z^o_t + z^a_t + z^v_t + p_t, \  h_t = \mathrm{GRU}(z_t, h_{t-1}).
\end{equation}
where $p_t$ is a temporal position embedding. The final progress estimate is produced by a small prediction head:
\begin{equation}
v_t = \sigma\big(\mathrm{MLP}([z^o_t, h_t, z_g])\big),
\end{equation}
where $z_g$ is the task-instruction embedding encoded by the base WAM and $\sigma$ is the sigmoid function. This design is expressive enough to capture execution dynamics while remaining lightweight compared to a full WAM inference call.

\subsection{AEP Module}
\label{sec:rule}
The progress estimate $v_t$ alone is not sufficient for execution control, because the same amount of progress has different physical meaning in various length tasks. 
For example, advancing by 0.05 over 10 steps is too slow for a 20-step task, but perfectly reasonable for a 200-step task. TempoWAM therefore compares progress rates rather than absolute progress increments.


Let $u_t$ denote the number of steps already executed when the $t$-th decision is made. We define the observed and the required progress rate as
\begin{equation}
\begin{aligned}
    \rho_t^{\mathrm cur} &= \frac{v_t - v_{t-1}}{n_{t}}, \\ 
    \rho_t^{\mathrm need} &= \max(\frac{1 - v_{t-1}}{\max(\bar T_{\mathrm ep} - u_t, n_t)}, \frac{0.2}{\bar T_{\mathrm ep}}),
\end{aligned}
\end{equation}
where $n_{t}$ is the candidate prefix length, and $\bar T_{\rm ep}$ is the average episode length of the task's demonstrations. $\rho_t^{\rm cur}$ measures how much progress each executed step actually advances, while $\rho_t^{\rm need}$ measures how much each remaining step needs to advance. 

The raw ratio
\begin{equation}
\tilde{r}_t = \rho_t^{\mathrm cur} / \rho_t^{\mathrm need}
\end{equation}
is therefore scale-free: values below 1 indicate that the remaining actions are advancing too slowly.

To make this ratio usable in practice, TempoWAM introduces a calibration factor $\kappa$ that aligns the monitor output with the execution state.

\paragraph{Offline calibration.} 
For each demonstration episode of a task, we run the monitor autoregressively and fit an episode-level factor $\kappa_{\mathrm{ep}}$ by least squares: 
\begin{equation}
\begin{aligned}
\kappa_{\mathrm{ep}} = \arg\min_{\kappa} \sum_u \left(\hat v_u - \kappa \frac{u}{\bar T_{\mathrm{ep}}}\right)^2,
\label{eq:kappa-offline}
\end{aligned}
\end{equation}
We then aggregate episodes by the median, $\kappa_{\mathrm task} = \operatorname{median}(\{\kappa_{\mathrm ep}\})$, which provides a robust initialization for each task.

\paragraph{Online adaptation.} 
At deployment time, the calibration factor is adapted online at two timescales. Within an episode, TempoWAM initializes $\kappa \leftarrow \kappa_{\rm task}$ and updates it by an EMA over the observed raw ratios:
\begin{equation}
    \kappa \;\leftarrow\; \lambda\,\kappa \;+\; (1-\lambda)\,\tilde{r}_t,
    \label{eq:ema}
\end{equation}
where $\lambda \in [0,1]$ is the EMA coefficient. This update compensates for the training-execution gap and tracks the current episode’s scale drift.

Across episodes, TempoWAM further adjusts a scaling factor $\delta'$ toward a target success rate $s^*$:
\begin{equation}
    \delta' \leftarrow \mathrm{clip}\!\left(\delta' - \eta\,\frac{\hat{s}
    - s^*}{s^*},\; \delta_{\min},\; \delta_{\max}\right),
    \label{eq:delta}
\end{equation}
where $\hat{s}$ is an EMA of recent episode success. When success runs below target, $\delta'$ grows and the protocol replans more frequently; when success exceeds target, $\delta'$ shrinks to improve efficiency. The two factors act as the final calibration $\kappa_{\mathrm final} = \kappa \cdot \delta'$, giving the decision rule
\begin{equation}
    r_t \;=\; \tilde{r}_t \,/\, \kappa_{\mathrm final}, \ 
    \text{continue if } r_t \geq 1, \;\; \text{replan if } r_t < 1.
    \label{eq:rule}
\end{equation}
For tasks without demonstrations, the offline $\kappa_{\rm task}$ is unavailable. In that case, TempoWAM initializes $\kappa=1$ and relies on the two-timescale online adaptation to gradually compensate for the missing calibration.

\begin{table*}[t]
\centering
\small
\setlength{\tabcolsep}{4pt}
\begin{tabular}{ll ccc ccc ccc ccc ccc}
\toprule
\multirow{2}{*}{\textbf{Task}} & \multirow{2}{*}{\textbf{Env.}} & \multicolumn{3}{c}{\textbf{Baseline-24}} & \multicolumn{3}{c}{\textbf{Baseline-12}} & \multicolumn{3}{c}{\textbf{TempoWAM}} & \multicolumn{3}{c}{\textbf{Auto-Horizon}} & \multicolumn{3}{c}{\textbf{AAC}} \\
\cmidrule(lr){3-5} \cmidrule(lr){6-8} \cmidrule(lr){9-11} \cmidrule(lr){12-14} \cmidrule(lr){15-17}
 &  & SR & Calls & Steps & SR & Calls & Steps & SR & Calls & Steps & SR & Calls & Steps & SR & Calls & Steps \\
\midrule
\multirow{2}{*}{open\_microwave} 
  & Clean & 53 & 19.04 & 441.5 & 51 & 25.71 & 303.5 & \textbf{83} & 27.55 & 352.3 & 52 & 35.40 & 488.4 & 45 & 109.4 & 384.0 \\
  & Rand. & 41 & 21.12 & 492.3 & 39 & 43.67 & 518.6 & \textbf{56} & 34.04 & 428.6 & 27 & 42.74 & 591.2 & 43 & 112.3 & 419.5 \\
\multirow{2}{*}{turn\_switch} 
  & Clean & 60 & 3.30 & 71.3 & 53 & 6.19 & 69.3 & \textbf{67} & 5.15 & 70.6 & 58 & 4.50 & 57.7 & 51 & 23.8 & 83.0 \\
  & Rand. & 64 & 3.34 & 72.8 & 67 & 6.63 & 74.6 & \textbf{74} & 5.30 & 69.6 & 69 & 5.10 & 65.4 & 55 & 23.9 & 78.3 \\
\midrule
\multirow{2}{*}{click\_bell} 
  & Clean & \textbf{100} & 3.00 & 56.9 & \textbf{100} & 5.26 & 58.2 & \textbf{100} & 2.04 & 60.6 & \textbf{100} & 4.26 & 53.5 & 82 & 35.0 & 93.9 \\
  & Rand. & \textbf{100} & 3.00 & 56.5 & \textbf{100} & 5.28 & 58.0 & \textbf{100} & 2.02 & 60.8 & 99 & 4.34 & 54.0 & 78 & 30.1 & 92.1 \\
\multirow{2}{*}{press\_stapler} 
  & Clean & 93 & 4.11 & 86.7 & 94 & 6.66 & 76.8 & \textbf{97} & 3.48 & 84.8 & 94 & 5.37 & 71.6 & 82 & 30.7 & 98.2 \\
  & Rand. & 96 & 4.01 & 83.8 & \textbf{98} & 6.67 & 76.9 & 96 & 3.55 & 82.0 & 97 & 5.42 & 72.4 & 93 & 29.3 & 93.8 \\
\bottomrule
\end{tabular}
\caption{Representative RoboTwin task results. TempoWAM improves success on difficult tasks and reduces calls on easy ones.}
\label{tab:robotwin-tasks}
\end{table*}

\subsection{Training Objective}\label{sec:training}
TempoWAM is trained on the same demonstration trajectories used by the backbone WAM. For a trajectory of length $T_{\mathrm{ep}}$, the supervision target at step $u$ is
$y_u = {u} / {T_{\mathrm{ep}}}$.
This label provides a monotonic signal of task progress and is sufficient for learning an execution-oriented monitor.

\paragraph{Progress prediction loss.}
We train the monitor with a binary cross-entropy objective on the soft progress label:
\begin{equation}
\begin{aligned}
\mathcal{L}_{\rm BCE} &= -\frac{1}{|B|} \sum_{(i,u)\in B} m_u^{(i)} \\
&\left[
y_u^{(i)} \log v_u^{(i)}
+
\left(1-y_u^{(i)}\right)\log\left(1-v_u^{(i)}\right)
\right],
\end{aligned}
\end{equation}
where $B$ is a mini-batch, and $m_u^{(i)}$ is a binary mask that excludes padded timesteps.

\paragraph{Smoothness regularization. }
Because the execution rule depends on differences between consecutive progress estimates, we additionally encourage the predicted progress to be smooth over time:
\begin{equation}
\begin{aligned}
\mathcal{L}_{\text{smooth}} &= \frac{1}{|B|} \sum_{(i,u) \in B} m_u^{(i)} m_{u+1}^{(i)}  \\
&\left|
({v}_{u+1}^{(i)} - {v}_u^{(i)}) - (y_{u+1}^{(i)} - y_u^{(i)})
\right|.
\end{aligned}
\end{equation}
This term reduces abrupt oscillations in the progress trajectory and makes the decision rule more stable.

\paragraph{History consistency regularization.}
To make the monitor less sensitive to brittle frame-level details, we apply history consistency regularization (HCR). With probability $p_{\mathrm{hcr}}$, we mask a subset of observations in the input window by setting them to zero, obtain a masked prediction $v^{\mathrm{mask}}_u$, and penalize the discrepancy between the masked and original predictions:
\begin{equation}
\mathcal{L}_{\text{hcr}} = \frac{1}{|B|}\sum_{(i,u)\in B}m_u^{(i)}
\left({v}_u^{(i)} - {v}^{\text{mask},(i)}_u\right)^2.
\end{equation}
This regularizer encourages the progress monitor to rely more on execution history rather than transient visual noise.
The full objective becomes
\begin{equation}
\mathcal{L} = \mathcal{L}_{\text{BCE}} + \lambda_{\text{smooth}} \mathcal{L}_{\text{smooth}} + \lambda_{\text{hcr}} \mathcal{L}_{\text{hcr}}.
\end{equation}



     
\section{Experiments}
\begin{table*}[t]
\centering
\small
\begin{tabular}{l ccc ccc ccc ccc}
\toprule
\multirow{2}{*}{Method} & \multicolumn{3}{c}{\textbf{Baseline}} & \multicolumn{3}{c}{\textbf{TempoWAM}} & \multicolumn{3}{c}{\textbf{Auto-Horizon}} & \multicolumn{3}{c}{\textbf{AAC}} \\
\cmidrule(lr){2-4} \cmidrule(lr){5-7} \cmidrule(lr){8-10} \cmidrule(lr){11-13}
 & SR & Calls & Steps & SR & Calls & Steps & SR & Calls & Steps & SR & Calls & Steps \\
\midrule
libero\_10     & 95.00 & 24.82 & 243.8 & 95.60 & \textbf{20.73} & 246.1 & 93.80 & 34.78 & 237.8 & 94.60 & 31.94 & 244.6 \\
libero\_spatial& 97.00 & 10.52 & 100.7 & 96.60 & \textbf{8.15}  & 99.9  & 96.80 & 14.62 & 98.6 & 97.60 & 15.12 & 100.7 \\
libero\_object & 99.40 & 13.16 & 127.5 & 99.40 & \textbf{11.17} & 128.4 & 97.40 & 18.98 & 127.6 & 99.20 & 19.75 & 128.6 \\
libero\_goal   & 97.60 & 11.03 & 105.9 & 97.40 & \textbf{9.80}  & 103.5 & 95.20 & 15.18 & 102.7 & 96.80 & 15.35 & 106.9 \\
\midrule
\textbf{Average}& \textbf{97.25} & 14.88 & 144.5 & \textbf{97.25} & \textbf{12.46} & 144.5 & 95.80 & 20.89 & 141.7 & 97.05 & 20.54 & 145.2 \\
\bottomrule
\end{tabular}
\caption{LIBERO benchmark results. TempoWAM reduces the WAM calls while preserving success across all four suites.}
\label{tab:libero_main}
\end{table*}

\begin{table}[t]
\centering
\small
\setlength{\tabcolsep}{4pt}
\begin{tabular}{lcccc}
\toprule
\multirow{2}{*}{Task} & \multicolumn{2}{c}{Motus SR (\%) $\uparrow$} & \multicolumn{2}{c}{TempoWAM SR (\%) $\uparrow$} \\
\cmidrule(lr){2-3} \cmidrule(lr){4-5} 
& clean & random & clean & random \\
\midrule
move\_can\_pot & 35 & 75 & 94 & 87 \\
place\_mouse\_pad & 61 & 60 &  93 & 86 \\
\bottomrule
\end{tabular}
\caption{Results of TempoWAM attached to Motus without retraining. This plug-and-play result supports that TempoWAM can be transferred across different WAMs.}
\label{tab:motus}
\end{table}

\begin{figure*}[t]
    \centering
    \includegraphics[width=0.95\linewidth]{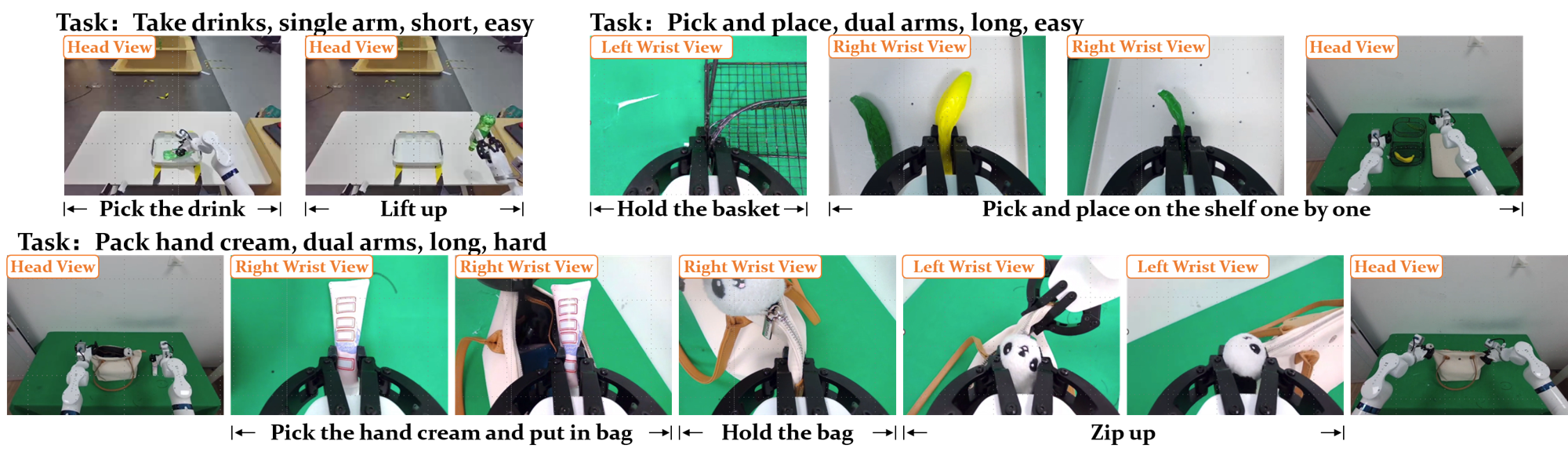}
    \caption{Real-robot task definitions and visualizations.}
    \label{fig:real-task}
\end{figure*}

\subsection{Experimental Setup}
\label{sec:expsetup}
\paragraph{Benchmarks and backbone.}
We evaluate TempoWAM using both simulation benchmarks RoboTwin 2.0~\cite{chen2025robotwin} and LIBERO~\cite{liu2023libero}), and real-world manipulation tasks. 
The primary backbone WAM is the frozen FastWAM~\cite{yuan2026fastwam}. To verify plug-and-play generalization, we additionally deploy TempoWAM to Motus~\cite{bi2025motus} without further training. 
On RoboTwin 2.0, we evaluate 50 tasks under clean and randomized settings with 100 trials per task. 
On LIBERO, we follow the standard protocol on 4 suites: LIBERO-Long, LIBERO-Spatial, LIBERO-Object, and LIBERO-Goal, with 50 trials per task. We report our real robot experimental results on three tasks, depicted in Fig.~\ref{fig:real-task}.

\paragraph{Baselines.}
Our primary baseline is the default fixed-horizon execution of FastWAM: replanning every 24 steps on RoboTwin and every 10 steps on LIBERO. On RoboTwin we additionally compare against replanning every 12 steps to test whether simply replanning more frequently can match adaptive execution. We further compare two representative adaptive-execution methods: Auto-Horizon~\cite{wang2026autohorizon}, which reads the predictive limit from action self-attention, and AAC~\cite{liang2026adaptive}, which selects the execution length from action entropy. For AAC we use the official implementation with $N{=}5$ samples per decision; its own ablation shows that increasing $N$ to 20 changes success by less than 1\%. Each AAC decision is counted as one inference call, although 5 samples are drawn in a single forward.

\paragraph{Metrics.}
We report success rate (SR, \%), diffusion inference calls and executed steps of successful episodes. Calls measure computational cost, as each call invokes one full WAM inference. Steps measure trajectory efficiency.

\paragraph{Implementation details.}
The monitor uses a GRU core, a GRU action encoder, and two-layer MLP projectors, totaling 2.27M parameters (0.038\% of the backbone FastWAM). It is trained for 20 epochs with $p_{\rm tf}{=}$0.5, $p_{\rm hcr}{=}$0.3, $\lambda_{\rm smooth}{=}$0.1, and $\lambda_{\rm hcr}{=}$0.05. At execution time, $n_p{=}$10 on RoboTwin and LIBERO, $n_p{=}$4 on real-world tasks. The online calibration uses EMA coefficient $\lambda{=}0.5$. The inter-episode scaling uses $\delta'_{\rm init}{=}0.8$, target success rate $s^*{=}0.95$, $\eta{=}0.05$, $[\delta_{\min},\delta_{\max}]{=}[0.4,1.0]$. Simulation experiments run on NVIDIA H20.

\subsection{Simulation Experiments}
\paragraph{Results on RoboTwin.}
Following the baseline execution cost, we group tasks whose average calls are $<10$ as short-horizon and the rest as long-horizon.
Table~\ref{tab:robotwin-grouped} shows that TempoWAM adapts in opposite directions on the two groups. On short-horizon tasks, it slightly improves or maintains success while reducing WAM calls. On long-horizon tasks, it replans earlier when chunks become unreliable, which improves success from 86.93\% to 88.47\% for clean ones and 84.47\% to 85.33\% for randomized ones, while also reducing executed steps. Simply reducing the execution horizon to 12 steps does not reproduce the improvement of TempoWAM: although it nearly doubles the number of calls, it still lowers success. This indicates that shorter horizons alone are not enough; what matters is when replanning is triggered.

\paragraph{Task-level analysis.}
Table~\ref{tab:robotwin-tasks} reports representative RoboTwin tasks. TempoWAM is especially effective on difficult tasks: on open\_microwave, success rises from 53\% to 83\% in clean setting and 41\% to 56\% under randomization, while significantly reducing executed steps. On smooth tasks it instead saves computation: on click\_bell, it preserves 100\% success while cutting calls from 3.00 to 2.04. The two adaptive baselines fail: Auto-Horizon replans excessively without recovering success (35.4 calls at 52\% success on open\_microwave), and AAC both replans the most (109.4 calls) and degrades success (45\%).

\begin{table}[t]
\centering
\small
\setlength{\tabcolsep}{3pt}
\begin{tabular}{l cc cc cc}
\toprule
\multirow{2}{*}{Method} & \multicolumn{2}{c}{\textbf{Take drinks}} & \multicolumn{2}{c}{\textbf{Pick and place}} & \multicolumn{2}{c}{\textbf{Pack hand cream}} \\
\cmidrule(lr){2-3} \cmidrule(lr){4-5} \cmidrule(lr){6-7} 
 & SR & Calls & SR & Calls & SR & Calls \\
\midrule
Baseline & 90.0 & 32.7 & 93.3 & 64.4 & 50.0 & 49.4 \\
TempoWAM & 90.0 & 23.9 & 96.7 & 68.7 & 63.3 & 57.2 \\
\bottomrule
\end{tabular}
\caption{Results on real robot results. The results show the same stage-aware behavior observed in simulation.}
\label{tab:real-task}
\end{table}

\begin{table}[t]
\centering
\small
\setlength{\tabcolsep}{5pt}
\begin{tabular}{@{}lcc@{}}
\toprule
 & RTX 4090 (ms) & Thor (ms) \\
\midrule
\multicolumn{3}{@{}l}{\emph{WAM inference call (ms)}} \\
PyTorch baseline                      & 62.15 & 181.23 \\
+ FP16 TensorRT                       & 27.02 & 84.17  \\
\quad + custom Blackwell kernels      &  --   & 78.91  \\
+ FP8 mixed precision                 & 20.36 & 53.21  \\
+ FP4 mixed precision                 &  --   & 47.17  \\
+ CUDA Graph (full stack)             &  --   & \textbf{42.31} \\
\midrule
\multicolumn{3}{@{}l}{\emph{RPM (\% of one optimized call)}} \\
with visual encoding                  & \multicolumn{2}{c}{3.54\%} \\
with shared features       & \multicolumn{2}{c}{\textbf{0.34\%}} \\
\bottomrule
\end{tabular}
\caption{Cost of one inference call and one RPM decision. `--' indicates stages not applicable on platform. }
\label{tab:overhead}
\end{table}

\paragraph{Results on LIBERO.}
Table~\ref{tab:libero_main} shows that TempoWAM preserves success on LIBERO while reducing WAM calls from 14.88 to 12.46 on average. Notably, on LIBERO-Long, which has the longest horizons, TempoWAM slightly improves success while cutting calls from 24.82 to 20.73. Auto-Horizon and AAC both require more calls than the fixed baseline without improving success.

\paragraph{Cross-backbone generalization.}
We additionally attach TempoWAM to Motus without retraining in Table~\ref{tab:motus}. TempoWAM improves success by up to \textbf{59 points} on move\_can\_pot. Note that Motus executes its full 16-step chunk by default, leaving no room to extend reuse on easy tasks; the gains here therefore come purely from earlier replanning on hard stages. This confirms that progress-driven adaptive execution is a property of the execution layer, not of a specific WAM architecture.

\begin{figure*}[t]
    \centering
    \includegraphics[width=0.98\linewidth]{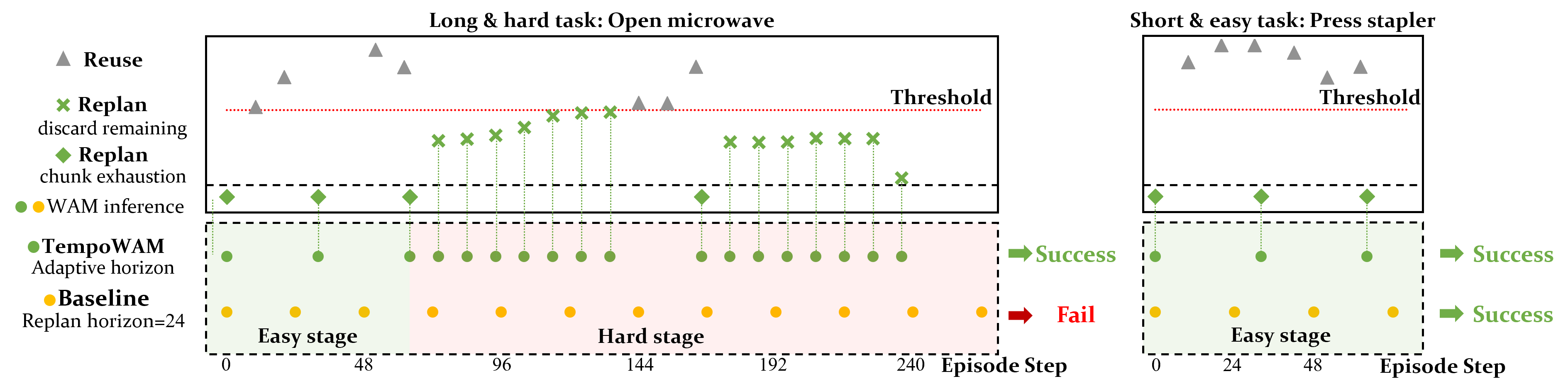}
    \caption{\textbf{Adaptive execution behavior on two RoboTwin episodes.} Top: the calibrated ratio $r_t$ and the decision threshold. Bottom: execution timelines of TempoWAM and the fixed-horizon baseline, with easy and hard stages shaded. Left: on open\_microwave, TempoWAM discards the unreliable chunk early at the hard stage and succeeds, while the baseline fails; right: on press\_stapler, the chunk remains reliable and TempoWAM reuses it fully to reduce WAM calls.}
    \label{fig:trace}
\end{figure*}

\begin{figure}[t]
    \centering
    \includegraphics[width=\linewidth]{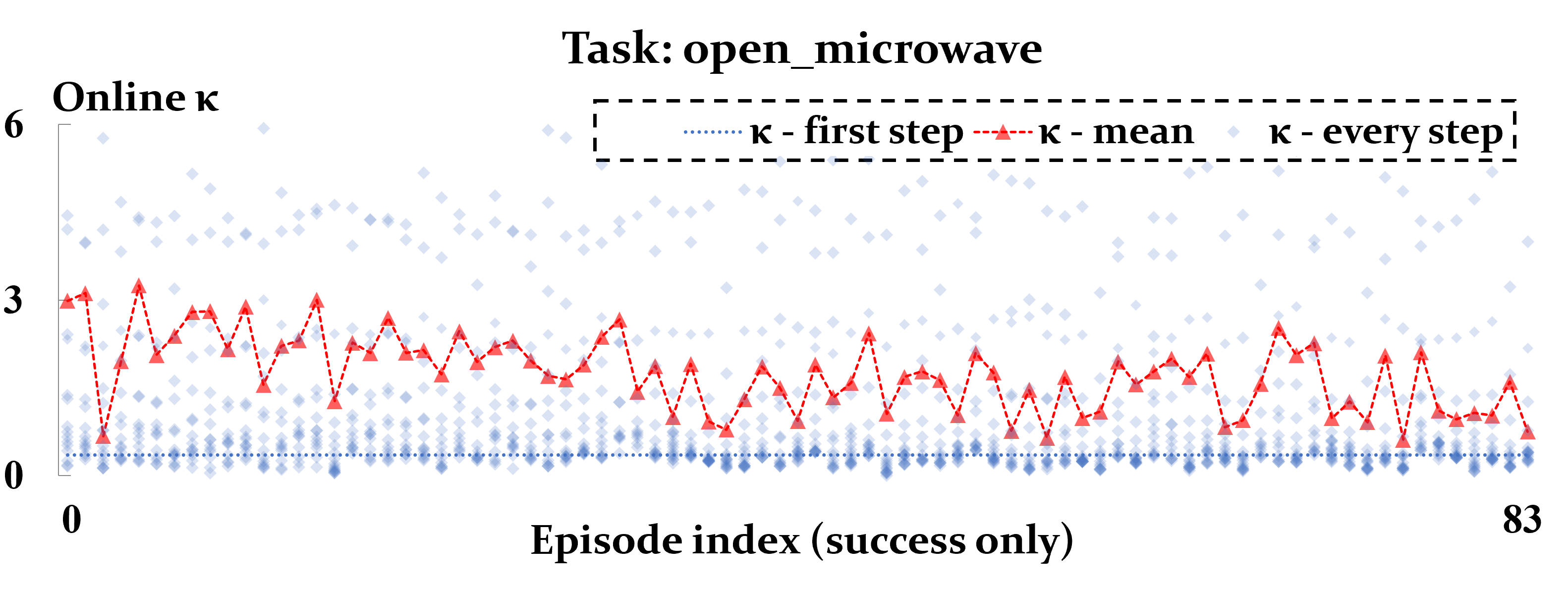}
    \caption{\textbf{Intra-episode calibration on open\_microwave}, with the inter-episode update of $\delta'$ disabled. Each episode starts with $\kappa$ at its offline initialization and rises within the episode, showing that the online EMA alone recognizes increasing task difficulty and tightens the protocol accordingly.}
    \label{fig:kappa}
\end{figure}

\subsection{Real-World Experiments and Efficiency}

We deploy TempoWAM on a dual-arm robotic platform, and evaluate three tasks with increasing difficulty: take drinks (short, easy), pick and place (long, easy), pack hand cream (long, hard), with 30 trials per task, as shown in Fig.~\ref{fig:real-task}.

\paragraph{Results.} 
As shown in Table~\ref{tab:real-task}, TempoWAM preserves success on take drinks while reducing calls from 32.7 to 23.9 ($-26.9\%$), improves pick and place from 93.3\% to 96.7\% with a modest increase in calls, on pack hand cream, the hardest task, success improves from 50.0\% to 63.3\%. These real-world results are consistent with results in  simulation: on easy tasks, TempoWAM reuses reliable chunks more aggressively; on difficult tasks, it replans earlier to recover success.

\paragraph{Deployment cost. }
Calls are the appropriate cost metric because WAM inference dominates episode wall-time. As shown in Table~\ref{tab:overhead}, our deployment optimization reduces one WAM call from 181.2\,ms to 42.3\,ms on Thor ($4.3\times$) and to 20.4\,ms on RTX~4090 ($3.1\times$).
Against this cost, the monitor is negligible: it adds 3.54\% of one call per decision with visual encoding, and only 0.34\% when reusing the visual features already computed by the WAM. Since each WAM call is more expensive than an RPM query, the monitor's cost is marginal relative to the replanning it saves or triggers.

\begin{table}[t]
\centering
\small
\begin{tabular}{lccc}
\toprule
\multirow{2}{*}{Task} & \multicolumn{2}{c}{Success (\%, clean)} & \multirow{2}{*}{Final $\delta'$} \\
\cmidrule(lr){2-3}
& Baseline & TempoWAM & \\
\midrule
click\_bell & 100 & 100 & 0.68 \\
click\_alarmclock & 100 & 100 & 0.68 \\
turn\_switch & 60 & 67 & 0.98 \\
open\_microwave & 53 & 83 & 0.94 \\
\bottomrule
\end{tabular}
\caption{Final inter-episode scaling factor $\delta'$ and success rates on representative RoboTwin tasks. Starting from $\delta'{=}0.8$ for every task, the success-gated update relaxes $\delta'$ on tasks above the target success rate and tightens it on tasks below.}
\label{tab:delta}
\end{table}

\subsection{Analysis and Ablations}
\label{sec:analysis}
\paragraph{Online calibration behavior.}
Fig.~\ref{fig:kappa} isolates the intra-episode calibration behavior by disabling the cross-episode update of $\delta'$. For successful episodes of the difficult task open\_microwave, $\kappa$ starts each episode at its offline initialization (first-step values) and consistently rises within the episode (episode-level means), while per-step raw ratios fluctuate widely. The rising trend indicates that the monitor's raw estimates grow increasingly optimistic as execution enters harder stages, and the online EMA absorbs this bias. It pulls the calibrated ratio down and triggering earlier replanning.

\paragraph{Success-gated adaptation matches task difficulty.}
Table~\ref{tab:delta} reports the final inter-episode factor $\delta'$ together with success rates, all tasks starting from the same initial value $\delta'{=}0.8$. The success-gated update moves $\delta'$ in the direction of each task's gap from the target success rate. On tasks the backbone already solves (\emph{click\_bell}, \emph{click\_alarmclock}, 100\%), $\delta'$ relaxes to 0.68, trading excess success for efficiency: calls on \emph{click\_bell} drop from 3.00 to 2.04 with success unchanged. On below-target tasks, $\delta'$ tightens toward 1 (0.98 on \emph{turn\_switch}, 0.94 on \emph{open\_microwave}), shifting the protocol toward earlier replanning and recovering 7 and 30 points of success. The controller thus steers every task toward the target success rate from its own starting point, without any per-task tuning.

\paragraph{Adaptive execution behavior.}
Fig.~\ref{fig:trace} visualizes two representative RoboTwin episodes and highlights how TempoWAM adapts execution online. On open\_microwave, the calibrated ratio drops below the threshold once the chunk enters the hard stage, so TempoWAM discards the remaining unreliable actions and replans early. The fixed-horizon baseline, by contrast, keeps executing the same chunk on schedule and eventually fails. On press\_stapler, the chunk stays reliable throughout the episode, so TempoWAM executes it fully and finishes with fewer calls. These two cases illustrate the key behavior of TempoWAM: it extends chunk reuse when the current execution remains reliable, and switches to replanning once the predicted advancement becomes insufficient.

\paragraph{Training Loss Ablations.}

\begin{table}[t]
\centering
\footnotesize
\begin{tabular}{lcc}
\toprule
Model & Success (\%) & Calls \\
\midrule
Full model & 97.25 & 12.46 \\
\quad teacher forcing $=1$ & 96.20 {(-1.05)} & 11.02 {(-1.44)} \\
\quad teacher forcing $=0$ & 96.65 {(-0.60)} & 12.32 {(-0.14)} \\
\quad w/o HCR & 96.65 {(-0.60)} & 12.39 {(-0.07)} \\
\quad w/o smoothness reg. & 96.65 {(-0.60)} & 12.01 {(-0.45)} \\
\quad LSTM core & 97.10 {(-0.15)} & 13.10 {(+0.64)} \\
\bottomrule
\end{tabular}
\caption{Component ablation on LIBERO.}
\label{tab:ablation}
\end{table}

Table~\ref{tab:ablation} ablates the monitor's training Loss on LIBERO. 
Smoothness and history-consistency regularization each contribute about 0.6 points of success. 
Teacher forcing shows a telling asymmetry: removing it (TF${=}0$) costs 0.6 points, while always using it (TF${=}1$) hurts more ($-1.05$ success, $-1.44$ calls). With TF${=}1$, the monitor is trained only on ground-truth history but must consume its own estimates at deployment; the errors in this feedback loop were never seen during training and are left uncorrected, degrading the execution decisions. The asymmetry confirms the exposure-bias motivation for probabilistic teacher forcing: TF${=}0$ destabilizes training, TF${=}1$ suffers train-test mismatch, and $p_{\mathrm{tf}}{=}0.5$ balances the two.
Additionally, replacing the GRU core with an LSTM slightly reduces success and increases calls. Since the LSTM increases the monitor's parameter count and compute cost, we keep the GRU core.

\paragraph{When Are Time-Based Progress Labels Valid?}
\label{app:time-labels}

\begin{figure}[t]
    \centering
    \includegraphics[width=0.48\linewidth]{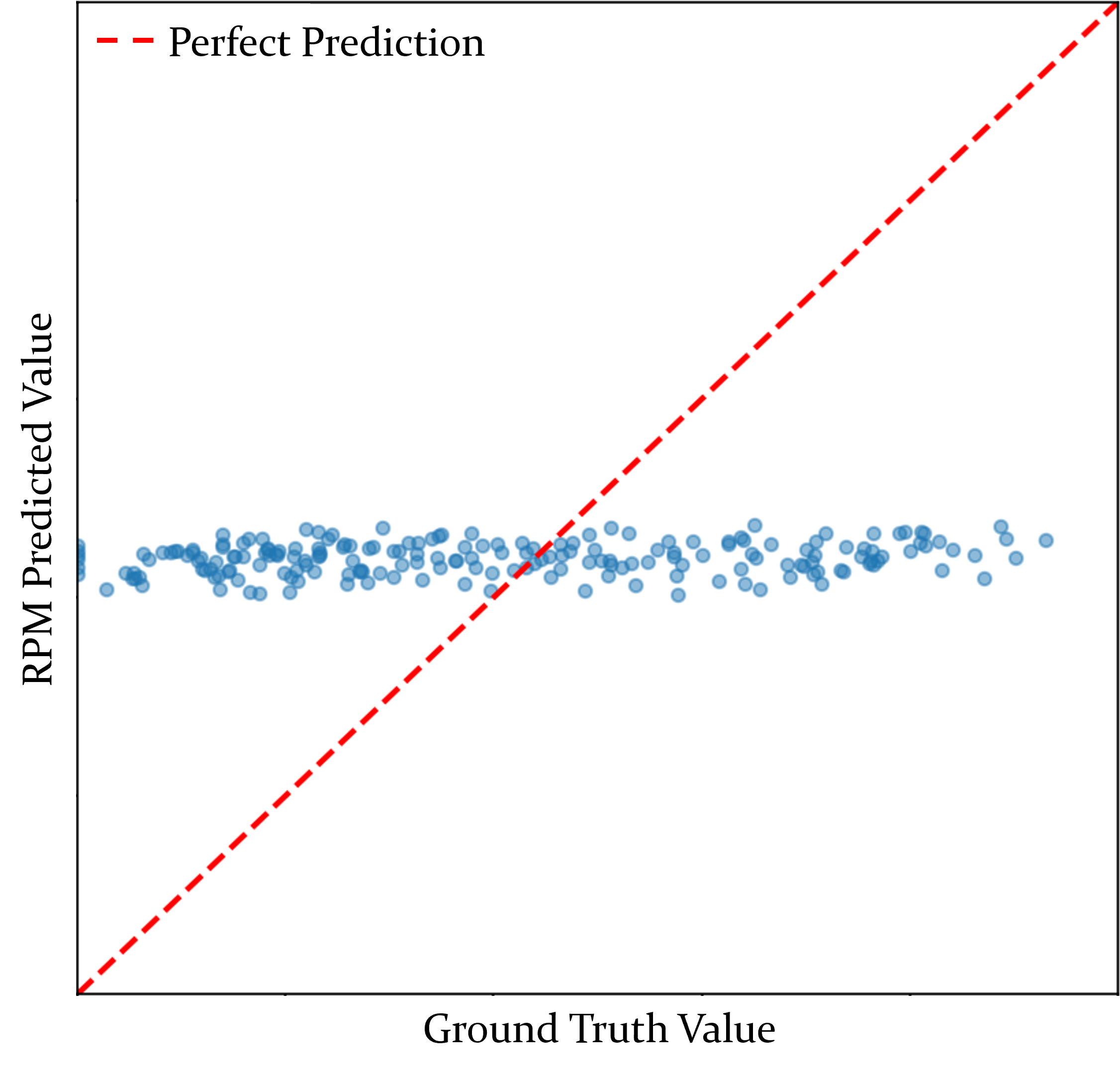}
    \caption{\textbf{Time-based labels fail on a mirrored sequence.} Ground-truth progress labels and RPM predictions on the palindromic take drinks construction. The predictions are unstructured and uncorrelated with the labels, showing that contradictory supervision from aliased states prevents any usable estimate.}
    \label{fig:time_label}
\end{figure}

We diagnose the validity of the time-based labels $y_u = u/T_{ep}$ with a controlled construction. We take demonstrations of take drinks--the short, easy real-robot task, on which the monitor trains and deploys normally--reverse each episode in time, and concatenate the reversed copy after the original. The resulting demonstrations sequence shows the robot completing the task and then visually un-doing it: the frame at normalized time $\tau$ reappears at $1-\tau$, and the two occurrences receive labels that sum to one. We then train the RPM on such sequences with the standard labeling rule.

The monitor learns nothing usable: its predictions are unstructured and show no correlation with the ground-truth labels (Fig.~\ref{fig:time_label}). The failure is structural. Under the mirrored construction, the observation-to-label mapping is one-to-many, so no monotonic function of the observation can fit the supervision, and the training signal is self-contradictory. Notably, although the step index is available to the monitor as an input, time-based labels is not learned: the contradictory visual supervision dominates the optimization.

Two conclusions follow. First, time-based labels are valid exactly when the observation sequence is progress-monotonic without aliasing: the scene at normalized time $\tau$ must be distinguishable from that at any $\tau' \neq \tau$. All simulation and real-robot tasks evaluated in this paper are non-repetitive sequences that satisfy this condition, while periodic or symmetric tasks violate it and require phase decomposition (labeling each segment from $0$ to $1$) or semantic progress supervision, which we leave to future work. Second, the failure itself indicates that the monitor's estimates are anchored to the visual state rather than to step counting---a step-counting monitor would fit the palindromic labels perfectly. On progress-monotonic tasks, the learned state-to-progress mapping aligns with task-relevant scene configurations (approaching, grasping, placing), so the estimates are visually grounded rather than purely temporal.

\section{Conclusion}
In this work, we have shown that the reliability of WAM action chunks is stage-dependent, so any fixed execution horizon is suboptimal. We further argue that task progress is a natural criterion for adaptive execution, and propose TempoWAM, a plug-and-play scheme that drives execution with estimated task progress through a lightweight recurrent monitor and a two-timescale calibrated protocol. Experiments on LIBERO, RoboTwin, and real-robot tasks on several backbone WAMs show that TempoWAM achieves adaptive execution: reusing chunks on easy stages and replanning earlier on difficult ones, consistently improving the efficiency-success trade-off. 
Future work includes learning progress supervision beyond time-based labels and extending the protocol to broader policies.

{
    \bibliographystyle{aaai2027}
    \bibliography{aaai2027}

\begin{thebibliography}{17}
\providecommand{\natexlab}[1]{#1}

\bibitem[{Bi et~al.(2026)Bi, Tan, Xie, Wang, Huang, Liu, Zhao, Feng, Xiang, Rong, Zhao, Liu, Su, Ma, Su, and Zhu}]{bi2025motus}
Bi, H.; Tan, H.; Xie, S.; Wang, Z.; Huang, S.; Liu, H.; Zhao, R.; Feng, Y.; Xiang, C.; Rong, Y.; Zhao, H.; Liu, H.; Su, Z.; Ma, L.; Su, H.; and Zhu, J. 2026.
\newblock Motus: A Unified Latent Action World Model.
\newblock In \emph{Conference on Computer Vision and Pattern Recognition (CVPR)}, 35101--35113.

\bibitem[{Cai et~al.(2026)Cai, Ling, Chu, Liu, Kang, Liang, Xu, Mao, Zhang, Yang, Ying, Zheng, and Mu}]{cai2026ahawam}
Cai, J.; Ling, L.; Chu, S.; Liu, Z.; Kang, J.; Liang, Z.; Xu, W.; Mao, Y.; Zhang, W.; Yang, X.; Ying, R.; Zheng, R.; and Mu, Y. 2026.
\newblock AHA-WAM: Asynchronous Horizon-Adaptive World-Action Modeling with Observation-Guided Context Routing.
\newblock arXiv:2606.09811.

\bibitem[{Chen et~al.(2026)Chen, Wang, Chen, Chen, Gao, Tang, Li, Liu, Yao, Li, Xu, and Yu}]{chen2026lawam}
Chen, J.; Wang, K.; Chen, K.; Chen, S.; Gao, F.; Tang, W.; Li, Z.; Liu, W.; Yao, Z.; Li, B.; Xu, Y.; and Yu, C. 2026.
\newblock LaWAM: Latent World Action Models for Efficient Dynamics-Aware Robot Policies.
\newblock arXiv:2606.15768.

\bibitem[{Chen et~al.(2025)Chen, Chen, Chen, Cai, Liu, Liang, Li, Lin, Ge, Gu et~al.}]{chen2025robotwin}
Chen, T.; Chen, Z.; Chen, B.; Cai, Z.; Liu, Y.; Liang, Q.; Li, Z.; Lin, X.; Ge, Y.; Gu, Z.; et~al. 2025.
\newblock RoboTwin 2.0: A Scalable Data Generator and Benchmark with Strong Domain Randomization for Robust Bimanual Robotic Manipulation.
\newblock arXiv:2506.18088.

\bibitem[{Du et~al.(2023)Du, Yang, Dai, Dai, Nachum, Tenenbaum, Schuurmans, and Abbeel}]{du2023advances}
Du, Y.; Yang, S.; Dai, B.; Dai, H.; Nachum, O.; Tenenbaum, J.; Schuurmans, D.; and Abbeel, P. 2023.
\newblock Learning Universal Policies via Text-Guided Video Generation.
\newblock In \emph{Advances in Neural Information Processing Systems (NeurIPS)}, volume~36, 9156--9172.

\bibitem[{Feng et~al.(2026)Feng, Cheng, Shi, Han, Yan, Hong, Tian, and Jiang}]{feng2026denoising}
Feng, X.; Cheng, Y.; Shi, C.; Han, B.; Yan, Y.; Hong, Y.; Tian, Z.; and Jiang, L. 2026.
\newblock Denoising Tells When to Replan: Denoising-Variance Adaptive Chunking for Flow-Based Robot Policies.
\newblock arXiv:2606.03847.

\bibitem[{Hu et~al.(2025)Hu, Guo, Wang, Chen, Wang, Zhang, Sreenath, Lu, and Chen}]{hu2025video}
Hu, Y.; Guo, Y.; Wang, P.; Chen, X.; Wang, Y.-J.; Zhang, J.; Sreenath, K.; Lu, C.; and Chen, J. 2025.
\newblock Video Prediction Policy: A Generalist Robot Policy with Predictive Visual Representations.
\newblock In \emph{International Conference on Machine Learning (ICML)}, volume 267, 24328--24346.

\bibitem[{Kim et~al.(2026)Kim, Gao, Lin, Lin, Ge, Lam, Liang, Song, Liu, Finn, and Gu}]{kim2026cosmos}
Kim, M.~J.; Gao, Y.; Lin, T.-Y.; Lin, Y.-C.; Ge, Y.; Lam, G.; Liang, P.; Song, S.; Liu, M.-Y.; Finn, C.; and Gu, J. 2026.
\newblock Cosmos Policy: Fine-Tuning Video Models for Visuomotor Control and Planning.
\newblock In \emph{International Conference on Learning Representations (ICLR)}.

\bibitem[{Liang et~al.(2026)Liang, Wang, Wang, Wang, Peng, Chen, Chua, and Vadakkepat}]{liang2026adaptive}
Liang, Y.; Wang, X.; Wang, K.; Wang, S.; Peng, X.; Chen, H.; Chua, D. K.~H.; and Vadakkepat, P. 2026.
\newblock Adaptive Action Chunking at Inference-time for Vision-Language-Action Models.
\newblock In \emph{Conference on Computer Vision and Pattern Recognition (CVPR)}, 20802--20811.

\bibitem[{Liu et~al.(2023)Liu, Zhu, Gao, Feng, Liu, Zhu, and Stone}]{liu2023libero}
Liu, B.; Zhu, Y.; Gao, C.; Feng, Y.; Liu, Q.; Zhu, Y.; and Stone, P. 2023.
\newblock LIBERO: Benchmarking Knowledge Transfer for Lifelong Robot Learning.
\newblock arXiv:2306.03310.

\bibitem[{Liu et~al.(2025)Liu, Hamid, Xie, Lee, Du, and Finn}]{liu2024bid}
Liu, Y.; Hamid, J.~I.; Xie, A.; Lee, Y.; Du, M.; and Finn, C. 2025.
\newblock Bidirectional Decoding: Improving Action Chunking via Guided Test-Time Sampling.
\newblock In \emph{International Conference on Learning Representations (ICLR)}.

\bibitem[{Nie et~al.(2026)Nie, Li, Zhang, Lao, Liu, Zhang, and Huang}]{nie2026pace}
Nie, J.; Li, J.; Zhang, J.; Lao, J.; Liu, C.; Zhang, T.; and Huang, S. 2026.
\newblock PACE: Phase-Aware Chunk Execution for Robot Policies with Action Chunking.
\newblock arXiv:2606.00537.

\bibitem[{Wang et~al.(2026{\natexlab{a}})Wang, Zhang, Yan, Kompella, and Liu}]{wang2026autohorizon}
Wang, H.; Zhang, G.; Yan, Y.; Kompella, R.~R.; and Liu, G. 2026{\natexlab{a}}.
\newblock VLA Knows Its Limits: Adaptive Execution Horizons for Robot Policies.
\newblock In \emph{European Conference on Computer Vision (ECCV)}.

\bibitem[{Wang et~al.(2026{\natexlab{b}})Wang, Zhang, Lin, Luo, Wang, Wang, and Qi}]{wang2026trust}
Wang, R.; Zhang, Y.; Lin, J.; Luo, K.; Wang, J.; Wang, Z.; and Qi, X. 2026{\natexlab{b}}.
\newblock When to Trust Imagination: Adaptive Action Execution for World Action Models.
\newblock arXiv:2605.06222.

\bibitem[{Yan et~al.(2026)Yan, Li, Yang, and Mu}]{yan2026progressvla}
Yan, H.; Li, Q.; Yang, J.; and Mu, Y. 2026.
\newblock ProgressVLA: Progress-Guided Diffusion Policy for Vision-Language Robotic Manipulation.
\newblock arXiv:2603.27670.

\bibitem[{Yuan et~al.(2026)Yuan, Dong, Liu, and Zhao}]{yuan2026fastwam}
Yuan, T.; Dong, Z.; Liu, Y.; and Zhao, H. 2026.
\newblock Fast-WAM: Do World Action Models Need Test-time Future Imagination?
\newblock arXiv:2603.16666.

\bibitem[{Zhao et~al.(2026)Zhao, Bogdanovic, Sohal, Tao, Darvish, Aspuru-Guzik, Shkurti, and Garg}]{zhao2026dynamic}
Zhao, Y.; Bogdanovic, M.; Sohal, A.; Tao, L.; Darvish, K.; Aspuru-Guzik, A.; Shkurti, F.; and Garg, A. 2026.
\newblock Dynamic Execution Horizon Prediction for Chunk-based Robot Policies.
\newblock arXiv:2606.11408.

\end{thebibliography}
}



\end{document}